%% file: main.tex
\documentclass[10pt,twocolumn,letterpaper]{article}

\usepackage{cvpr}              
\input{preamble}
\definecolor{cvprblue}{rgb}{0.21,0.49,0.74}
\usepackage[pagebackref,breaklinks,colorlinks,allcolors=cvprblue]{hyperref}
\usepackage{comment}
\usepackage{graphicx}
\usepackage{arydshln}
\usepackage{algorithm}
\usepackage{algorithmic}
\usepackage[table]{xcolor} 
\usepackage{colortbl}      

\usepackage{pifont}
\usepackage{makecell}
\def\paperID{*****} 
\def\confName{CVPR}
\def\confYear{2026}

\title{CaricHarmony: Contrastive Diffusion Paths for Identity-Preserving Caricature Synthesis}

\author{Dongyu Wang \qquad Dar-Yen Chen \qquad Yi-Zhe Song\\
  SketchX, CVSSP, University of Surrey\\
  {\tt\small \{dongyu.wang, d.chen, y.song\}@surrey.ac.uk}\\
  {\tt\small \url{https://dongyuuw.github.io/CaricHarmony/}}\\
}

\begin{document}
\twocolumn[{
\renewcommand\twocolumn[1][]{#1}%
\maketitle

\thispagestyle{empty}
\begin{center}
    \centering
    \captionsetup{type=figure}
    \includegraphics[width=\linewidth]{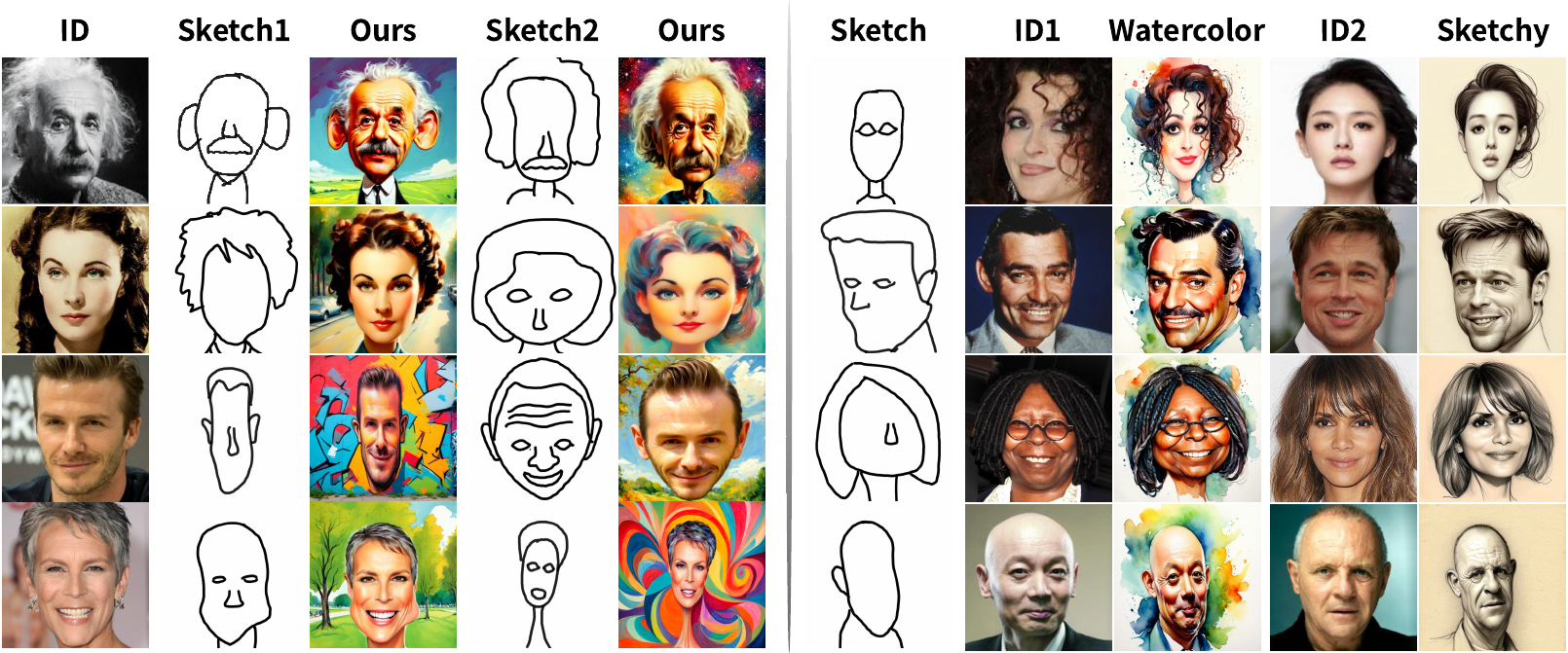}
    \vspace{-6mm}
    \caption{\textbf{High-quality caricatures generated by our CaricHarmony.} \textbf{Left:} Results generated with the same identity and different sketches. \textbf{Right:} Results generated with the same sketch and different identities and specified styles.}
    \label{fig:pilot}
\end{center}
}]

\begin{abstract}
Sketch-based caricature synthesis suffers from a fundamental failure mode: when identity and shape conditions are combined in diffusion models, they create destructive interference that causes inevitable collapse toward either bland portraits or unrecognizable distortions. We identify the root cause as \emph{condition signal contamination} -- competing probability distributions in the denoising trajectory that make balanced generation impossible. We present CaricHarmony, the first training-free method that explicitly resolves this contamination through parallel uncontaminated diffusion paths. During inference, we maintain three paths: $\mathcal{P}^{\mathrm{i}}$ (pure identity), $\mathcal{P}^{\mathrm{s}}$ (pure shape), and $\mathcal{P}^{\mathrm{i+s}}$ (harmonized output). Novel energy functions operating on cross-attention features provide gradient guidance that steers $\mathcal{P}^{\mathrm{i+s}}$ toward optimal balance: $\mathcal{E}_{\mathrm{shape}}$ ensures sketch fidelity through layout and semantic alignment, while $\mathcal{E}_{\mathrm{id}}$ employs token-level correspondence matching robust to extreme distortions. Unlike DemoCaricature requiring 70 seconds per-identity fine-tuning or CaricatureBooth constrained to Bezier curves, CaricHarmony accepts any sketch format and generates in under 16 seconds. Experiments demonstrate state-of-the-art performance: 0.8615 shape CLIP score (vs. 0.8450) under comparable identity consistency score, with 7.81 overall user preference score (vs. 6.06). Our method fundamentally reconceptualizes the ID-shape conflict as conditioning signal contamination for diffusion models, enabling unprecedented creative control while preserving recognition.
\end{abstract}

\section{Introduction}
\label{sec:intro}

Caricature synthesis demands a paradox: extreme shape distortions that somehow preserve identity. Artists instinctively navigate this tension, but computational methods systematically fail. When given a sketch with exaggerated features -- a grotesquely elongated nose, impossibly tiny eyes -- existing approaches either ignore these creative choices to preserve identity, or follow them faithfully at the cost of recognition. This fundamental \emph{ID-shape conflict} isn't just an implementation detail; it's the core challenge that has prevented sketch-based caricature synthesis from achieving its artistic potential.

We identify the root cause: \textit{condition signal contamination}. When identity conditions $C_{\mathrm{id}}$ and shape conditions $C_{\mathrm{s}}$ are combined in diffusion models, they create destructive interference in the denoising trajectory. The model oscillates between competing probability distributions -- regions near pure identity preservation and pure shape following exhibit higher densities than the balanced intermediate region, causing inevitable collapse toward one extreme. Prior methods \cite{DemoCaricature, CaricatureBooth} fundamentally ignore this problem, attempting means of balancing identity and shape without addressing the underlying contamination.

Table~\ref{tab:characteristics} reveals a critical gap in the field. Early synthesis methods~\cite{CariGAN, WarpGAN, AutoToon, StyleCariGAN} lack controllability entirely, performing automatic distortions that ignore user intent. Recent controllable approaches face crippling limitations: DemoCaricature~\cite{DemoCaricature} requires expensive per-sample fine-tuning (70 seconds per identity), making it impractical for interactive creation. CaricatureBooth~\cite{CaricatureBooth} constrains creativity to Bezier curves, requiring large-scale pre-training on synthetic data. Critically, \textit{neither addresses the core signal contamination issue} -- they simply shift where the ID-shape conflict manifests.

We present CaricHarmony, the first training-free solution that directly resolves condition contamination through \emph{contrastive feature alignment}. Our key insight transforms the problem: instead of mixing contaminated signals, we maintain three parallel diffusion paths during inference -- $\mathcal{P}^{\mathrm{i}}$ (pure identity), $\mathcal{P}^{\mathrm{s}}$ (pure shape), and $\mathcal{P}^{\mathrm{i+s}}$ (harmonized output). The uncontaminated reference paths $\mathcal{P}^{\mathrm{i}}$ and $\mathcal{P}^{\mathrm{s}}$ serve as guidance anchors, while specialized energy functions dynamically steer $\mathcal{P}^{\mathrm{i+s}}$ toward optimal balance through gradient guidance at each denoising step.

Our energy functions operate within cross-attention mechanisms to achieve precise control: $\mathcal{E}_{\mathrm{shape}}$ ensures sketch fidelity through layout alignment ($\mathcal{E}_{\mathrm{layout}}$) and semantic consistency ($\mathcal{E}_{\mathrm{sem}}$), while $\mathcal{E}_{\mathrm{id}}$ employs novel token-level correspondence matching that adapts to shape distortions. Unlike conventional identity losses that fail on caricatures due to domain gap, our approach leverages intermediate diffusion features that inherently preserve layout information for localizing facial features. The compositional energy $\mathcal{E}_{\mathrm{b}} = \mathcal{E}_{\mathrm{shape}} + \mathcal{E}_{\mathrm{id}}$ provides continuous gradient guidance throughout denoising, preventing trajectory collapse.

CaricHarmony achieves breakthrough results: generating high-quality caricatures from extreme sketches in 16 seconds (4× faster than DemoCaricature), accepting \textit{any sketch format} without preprocessing (unlike CaricatureBooth's Bezier constraints), and requiring \textit{zero training or fine-tuning} while maintaining state-of-the-art quality. Extensive experiments demonstrate superior performance across CLIP-based shape fidelity metric  \cite{CLIP, DemoCaricature} (0.8615 vs. 0.8450) and aesthetic quality metrics (ImageReward \cite{ImageReward} 0.8509 vs. 0.4340 and PickScore \cite{PickScore} 0.2049 vs. 0.1094) with competitive ID consistency. Furthermore, user studies reveal a strong preference for our results, achieving an overall preference score of 7.81 (vs. 6.06).

Our contributions are: (i) \textit{First explicit solution to condition contamination}: We identify signal contamination as the root cause of ID-shape conflict and resolve it through parallel uncontaminated reference paths, fundamentally different from prior diffusion-based approaches. (ii) \textit{Novel cross-attention energy functions}: We introduce specialized alignment mechanisms that operate on intermediate diffusion features, including token-level correspondence matching that adapts to arbitrary shape distortions -- impossible with conventional identity losses. (iii) \textit{Practical democratization of caricature creation}: Our training-free framework eliminates all bottlenecks -- no pre-training, no per-sample optimization, no sketch format constraints -- making professional caricature synthesis accessible to anyone with a rough sketch.

\begin{table}[!htbp]
    \scriptsize
    \centering
    \setlength\tabcolsep{8pt}
    \begin{tabular}{ccccc}
    \toprule
           Methods          & Training-free & \makecell{Controllable \\ generation} & \makecell{Free-form \\ conditioning}  \\
    \midrule
    StyleCariGAN \cite{StyleCariGAN} & \color{red}\ding{55}    &   \color{red}\ding{55}  & \color{green!60!black}\ding{51}\\
    WarpGAN \cite{WarpGAN} & \color{red}\ding{55} & \color{red}\ding{55} &\color{green!60!black}\ding{51}\\
    AutoToon \cite{AutoToon} & \color{red}\ding{55} & \color{red}\ding{55}&\color{green!60!black}\ding{51}\\
    DemoCaricature \cite{DemoCaricature} & \color{red}\ding{55} & \color{green!60!black}\ding{51} &\color{green!60!black}\ding{51}\\
    CaricatureBooth \cite{CaricatureBooth} & \color{red}\color{red}\ding{55} & \color{green!60!black}\ding{51} &\color{red}\ding{55}\\
    \arrayrulecolor{gray}\midrule\arrayrulecolor{black}
    Ours     & \color{green!60!black}\ding{51}& \color{green!60!black}\ding{51} &\color{green!60!black}\ding{51}\\
    \bottomrule
    \end{tabular}
    \caption{Our model features user-friendly characteristics beyond dealing with the ID-shape conflict issue. Free-form conditioning: no additional requirements on shape conditions (if applicable) and ID conditions.}
    \label{tab:characteristics}
    \vspace{-3mm}
\end{table}

\section{Related Work}
\label{sec:related}
\noindent\textbf{Deep Caricature Synthesis.} Caricature synthesis aims at generating a stylized portrait of a human subject with artistic styles and shape exaggerations while preserving the identity \cite{Semantic-CariGAN, CariMe}. CariGANs \cite{CariGAN} utilizes GAN-based networks to generate stylized caricatures under the guidance of deformed facial landmarks. WarpGAN \cite{WarpGAN} eases the warping process by using automatically generated control points. Several subsequent works include diversifying styles and exaggeration types \cite{CariMe}, using SENet \cite{SENet} to generate warping fields that densely guide the warping process \cite{AutoToon}, and introducing shape exaggeration blocks \cite{StyleCariGAN} based on StyleGAN \cite{StyleGAN} to achieve shape deformation by manipulating coarse-scale feature maps. Semantics CariGAN \cite{Semantic-CariGAN} proposes using facial parsing maps to guide the shape transformation in the output. For works realizing the controllability of the shape exaggeration, DemoCaricature \cite{DemoCaricature} first proposes using free-hand sketches as a flexible and expressive medium to guide the shape deformation process. It encapsulates the identity information into textual embeddings and the key-value pathway of cross-attention blocks through per-sample tuning, but lacks consideration of its potential conflict with the shape information, limiting its performance on exaggerated free-hand sketches. CaricatureBooth \cite{CaricatureBooth} represents a sketch with a fixed number of Bezier curves \cite{Bezier} and performs TPS deformation \cite{TPS} to generate synthetic data for training. However, it severely constrains the format of sketches and limits the diversity and creativity of shape conditions. 

In contrast to prior works, our approach explicitly addresses the fundamental ID-shape conflict through a novel training-free framework that balances identity preservation and shape exaggeration, overcoming the limitations observed in DemoCaricature. Furthermore, our method accepts free-hand sketches in their natural form, preserving the full expressive potential and creativity of sketch-based control while ensuring robustness to diverse and exaggerated shape conditions.

\noindent\textbf{T2I Personalization.} The goal of T2I personalization is to customize T2I models \cite{Imagen, SD, SDXL, flux} to consider concepts specified by a set of reference images while maintaining their original generalizability. Textual Inversion \cite{Textual-Inversion} finds new pseudo-word embeddings to represent the semantic features of referred subjects. DreamBooth \cite{DreamBooth} fine-tunes pretrained Stable Diffusion \cite{SD} and Imagen \cite{Imagen} models to link each specified subject with a unique representative textual identifier. For parameter-efficient fine-tuning, Custom Diffusion \cite{Custom-Diffusion} only fine-tunes cross-attention layers, while Perfusion \cite{Perfusion} presents a Rank-One Model Editing (ROME) method that optimizes only the value-pathway in cross-attention layers. For multi-subject optimization, LatexBlend \cite{LatexBlend} represents multiple concepts in a latent textual space, enabling multi-concept generation by retrieving concepts from a bank that can be freely combined without tuning. LoRACLR \cite{LoRACLR} merges multiple pretrained LoRAs into a unified model using contrastive learning to align multiple concepts while preventing interference. For fast personalization without per-sample tuning, IP-Adapter \cite{IP-Adapter} trains lightweight and decoupled cross-attention blocks to embed image prompts into the model in parallel with textual prompts. InstantBooth \cite{InstantBooth} achieves a tuning-free method by learning a network that can map reference images into aligned textual embeddings during inference. PuLID \cite{pulid} introduces a tuning-free approach that can output high-fidelity personalized images while mitigating the interruption to the behavior of the original model through contrastive alignment. Imagine yourself \cite{Imagine-yourself} proposes a parallel attention architecture with three text encoders and one vision encoder for high-fidelity identity preservation and text alignment. Flux already knows \cite{flux-already-knows} eliminates the need for pre-training and frames subject-driven generation as grid-based image completion with mosaic layouts by utilizing the generalizability of Flux models \cite{flux}.

\section{Revisit Text-to-Image Diffusion Models}

\label{sec:revisit}
\subsection{Diffusion Models}
Diffusion models \cite{DDPM, SD, SDXL} synthesize images from Gaussian noise by iterating the denoising process. They consist of two key components: a forward diffusion process and a reverse denoising process. The forward process corrupts a data sample $z_{0}$ by iteratively adding Gaussian noise $\epsilon \sim \mathcal{N}(0, \mathbf{I})$ over $T$ time-steps, producing noisy data $z_{T}$.  The denoising process recovers $z_{0}$ from $z_{T}$ by iteratively estimating and removing noise using a trained network $\epsilon_{\theta}(\hat{z}_{t}, t)$, which takes as input the predicted noisy latent $\hat{z}_{t}$ and its corresponding time-step $t$ \cite{DDPM, SD}. 
Typically, $\epsilon_{\theta}(\cdot)$ follows a UNET \cite{UNET} architecture and incorporates a series of residual blocks \cite{residual}, self-attention blocks, and cross-attention blocks \cite{Transformer}. 

From the continuous perspective \cite{Diffusion-SDE} of diffusion models, $\epsilon_{\theta}(\hat{z}_{t}, t)$ can be explained as a scaled version of the score function $\nabla_{\hat{z}_{t}} \log q(\hat{z}_{t})$ that samples from the probability distribution $q(\hat{z}_{t})$ \cite{score-based-model} through Langevin dynamics \cite{Langevin, Song-Langevin}. When the generative process is combined with external conditions $C$, the corresponding score function $\nabla_{\hat{z}_{t}} \log q(\hat{z}_{t}|C)$ can be decomposed into:
\begin{small}
\begin{equation}
\nabla_{\hat{z}_{t}} \log \frac{q(\hat{z}_{t})q(C|\hat{z}_{t})}{q(C)}= \nabla_{\hat{z}_{t}}\log q(\hat{z}_{t}) + \nabla_{\hat{z}_{t}}\log q(C|\hat{z}_{t}), 
\end{equation}
\end{small}
where the first term $\nabla_{\hat{z}_{t}}\log q(\hat{z}_{t})$ corresponds to an unconditional denoiser $\epsilon_{\theta}(\hat{z}_{t}, t)$, and the second term is the gradient guidance that can be generated by an energy function $-\mathcal{E}(\hat{z}_{t};t, C)=\log q(C|\hat{z}_{t})$. 

By designing energy functions for different generation objectives, such as text-conditioned generation \cite{CG}, sketch-guided generation \cite{sketch-guided-generation}, mask-guided generation \cite{mask-guided-generation}, and image editing \cite{score-based-image-editing}, the model can generate outputs that conform to specified conditions. Inspired by the success of prior works, we address the ID-shape conflict in caricature synthesis by designing tailored energy functions that explicitly guide noisy latent toward achieving an optimal trade-off. 

\subsection{IP-Adapter and PuLID}
IP-Adapter \cite{IP-Adapter} introduces an effective approach to integrate ID features, denoted as $C_{\mathrm{id}}$, into the model through the cross-attention mechanism  \cite{Transformer, SD} in parallel with textual features $C_{\mathrm{txt}}$. There are two linear layers $W_{K}^{\mathrm{id}}$ and $W_{V}^{\mathrm{id}}$ mapping $C_{\mathrm{id}}$ into key ($K_{\mathrm{id}}$) and value ($V_{\mathrm{id}}$) matrices. The 
cross attention is then calculated as follows: 
\begin{equation}
\begin{cases}
O=\mathrm{Attn}(Q, K_{\mathrm{id}}, V_{\mathrm{id}}) = \mathrm{Softmax}(\frac{Q{(K_{\mathrm{id}})^{T}}}{\sqrt{d}}) 
V_{\mathrm{id}} \\   \label{cross-attn}
Q=W_{Q}h_{t}, K_{\mathrm{id}}=W_{K}^{\mathrm{id}}C_{\mathrm{id}}, V_{\mathrm{id}}=W_{V}^{\mathrm{id}}C_{\mathrm{id}},
\end{cases}
\end{equation}
where $h_t\in R^{n \times d}$ represents hidden feature maps at time-step $t$ flattened along the height and width dimensions, $d$ denotes the latent dimension, and $n=h\times w$ represents the token number and is the product of the height $h$ and width $w$ of the hidden feature maps. The output $O$ is then added to the hidden feature maps. $C_{\mathrm{id}}$ is typically extracted using pretrained image encoders, such as the CLIP Image encoder \cite{CLIP} or face recognition models \cite{Arcface}. PuLID \cite{pulid} employs both encoders \cite{CLIP, Arcface} jointly and further emphasizes preserving the original model's behavior by carefully controlling how $C_{\mathrm{id}}$ is injected using contrastive alignment losses, thereby preventing unintended contamination of the generation process. 

\subsection{T2I Adapter}
T2I Adapter \cite{t2iadapter} presents a plug-and-play network that enables various spatial condition signals (e.g. sketches, edge maps, pose images, depth images) to guide the image generation process. The lightweight network consists of one convolution layer followed by two residual blocks at each scale. It maps input condition images into multi-scale condition features $C_{\mathrm{s}}$, which capture spatial information at different levels of granularity. These extracted features are then added to the intermediate feature maps of the encoder within the UNET denoiser at each scale.

\section{Methodology}

\begin{figure}[!tbp]
    \centering
    \includegraphics[width=\linewidth]{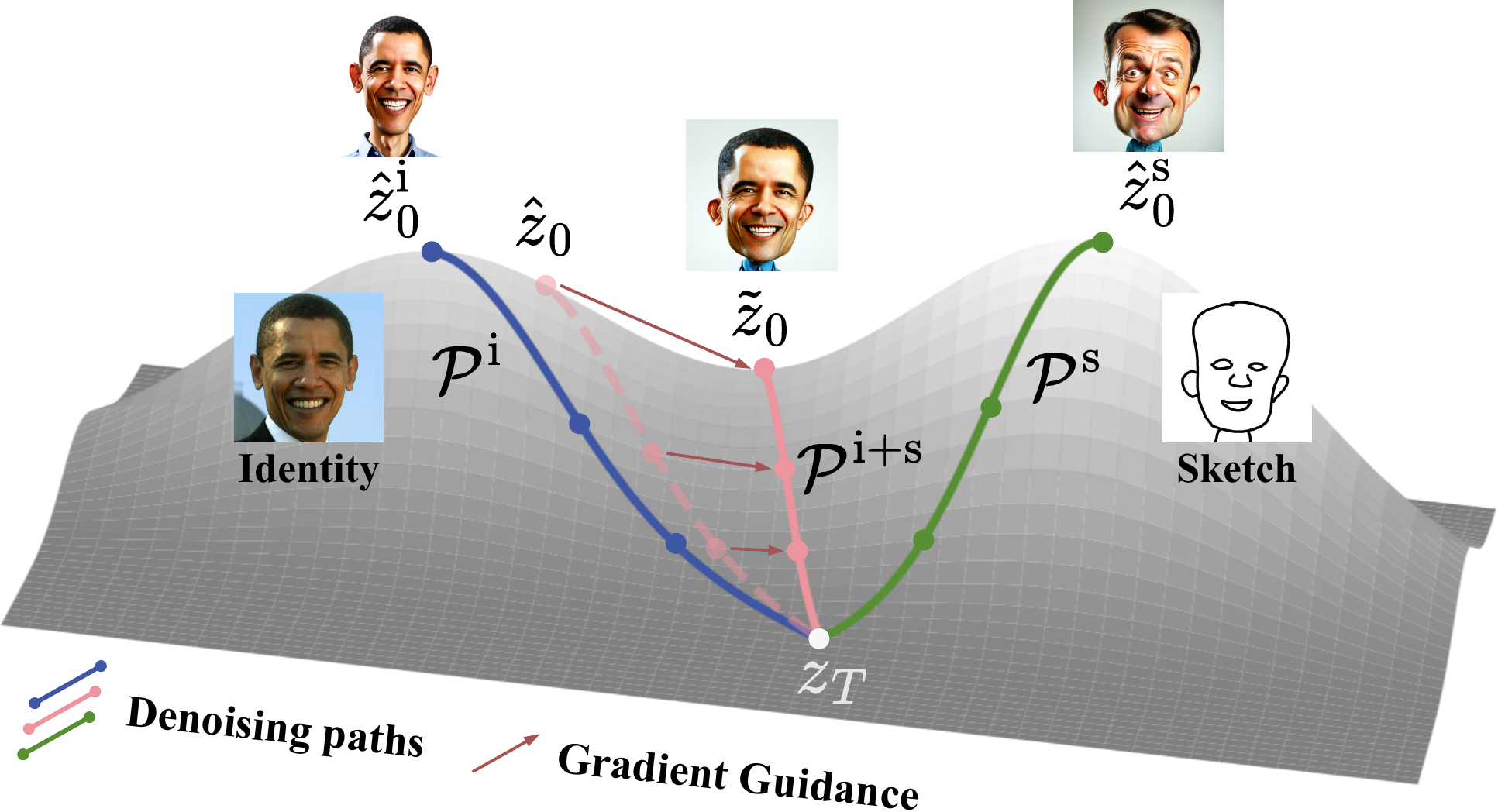}
    \vspace{-5mm}
    \caption{Dynamics of denoising paths in continuous sampling space of score-based diffusion \cite{Diffusion-SDE}. Brighter color indicates areas where the denoised data is densely distributed. In blue (green) path, only $C_{\mathrm{id}}$ ($C_{\mathrm{s}}$) is applied. Both conditions are applied in pink paths. The per-step gradient guidance from the energy function $\mathcal{E}_{\mathrm{b}}$ is represented by red arrows.}
    \label{fig:sampling}
    \vspace{-3mm}
\end{figure}

The ID–shape conflict refers to the tendency of the diffusion model to favor either identity preservation or shape exaggeration when both conditions are provided, while failing to balance them. As illustrated in \Cref{fig:sampling}, the ID-conditioned path $\mathcal{P}^{\mathrm{i}}$ naturally guides the denoising trajectory toward faithful identity preservation, whereas the shape-conditioned path $\mathcal{P}^{\mathrm{s}}$ drives it toward precise caricature geometry. When both conditions $C_{\mathrm{id}}$ and $C_{\mathrm{s}}$ are applied together, the ideal denoising trajectory should lie between these two extremes. However, since the regions near $\mathcal{P}^{\mathrm{i}}$ and $\mathcal{P}^{\mathrm{s}}$ exhibit higher probability densities than the intermediate region, the trajectory tends to collapse toward one side, overemphasizing either identity or shape.

To address this issue, we proposed CaricHarmony by constructing three denoising trajectories: the two contrastive paths $\mathcal{P}^{\mathrm{i}}$ and $\mathcal{P}^{\mathrm{s}}$, conditioned on $C_{\mathrm{id}}$ and $C_{\mathrm{s}}$ respectively, and the main path $\mathcal{P}^{\mathrm{i+s}}$ conditioned on both. For clarity, we denote intermediate features from $\mathcal{P}^{\mathrm{i}}$ and $\mathcal{P}^{\mathrm{s}}$ using superscripts ${\mathrm{i}}$ and ${\mathrm{s}}$, while features from $\mathcal{P}^{\mathrm{i+s}}$ are left without superscripts.

\begin{figure*}[tp]
    \centering
    \includegraphics[width=.95\linewidth]{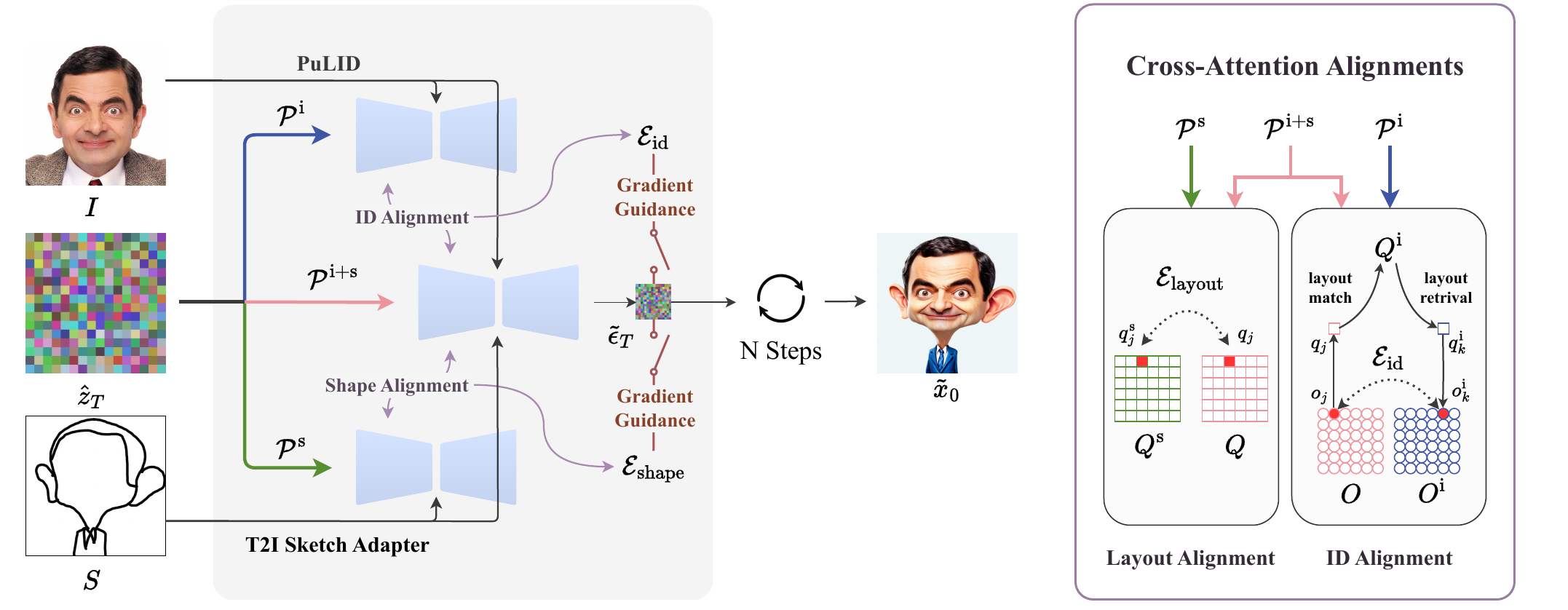}
    \vspace{-2mm}
    \caption{
    \textbf{Left: Inference pipeline of CaricHarmony}. The output path $\mathcal{P}^{\mathrm{i+s}}$ is jointly guided by two contrastive paths $\mathcal{P}^{\mathrm{s}}$ and $\mathcal{P}^{\mathrm{i}}$. $\mathcal{P}^{\mathrm{s}}$ is only fed the binary sketch map $S$ as shape conditions and guides the shape alignment through $\mathcal{E}_{\mathrm{shape}}$, which comprises $\mathcal{E}_{\mathrm{layout}}$ and $\mathcal{E}_{\mathrm{sem}}$. $\mathcal{P}^{\mathrm{i}}$ is only fed the ID conditions $I$ and guides the ID alignment through $\mathcal{E}_{\mathrm{id}}$.
    \textbf{Right: Cross-Attention Alignments.} Layout alignment obtains $\mathcal{E}_{\mathrm{layout}}$ by computing the token-wise differences between the query matrices from $\mathcal{P}^{\mathrm{i+s}}$ and $\mathcal{P}^{\mathrm{s}}$. ID alignment obtains $\mathcal{E}_{\mathrm{id}}$ by computing the token-wise differences between the ID cross attention outputs from $\mathcal{P}^{\mathrm{i+s}}$ and $\mathcal{P}^{\mathrm{i}}$. The associations between the output tokens for alignment are built by computing the similarities of corresponding query tokens.}
    \label{fig:archi}
    \vspace{-3mm}
\end{figure*}

We then impose inference-time cross-attention alignment guidance between the main path and each contrastive path, encouraging the main trajectory to stay within the region that balances identity and shape cues. The proposed gradient guidance mechanism explicitly steers the denoising direction away from biased extremes and toward the balanced region between $\mathcal{P}^{\mathrm{i}}$ and $\mathcal{P}^{\mathrm{s}}$, effectively resolving the ID–shape conflict without any additional training. The overall architecture is shown in \Cref{fig:archi}.

\subsection{Shape Alignments}
\label{sec:shape}

Without identity conditions, $\mathcal{P}^{\mathrm{s}}$ fully encodes shape exaggeration according to the sketch. To ensure that $\mathcal{P}^{\mathrm{i+s}}$ retains this capability, we align its intermediate cross-attention features with those from $\mathcal{P}^{\mathrm{s}}$ during inference.

Inspired by PuLID \cite{pulid}, we use the query features $Q^{\mathrm{s}}$ from each cross-attention block as shape layout target, since they reflect how shape-related textual and sketch conditions are spatially grounded. By aligning $Q$ with $Q^{\mathrm{s}}$, the model can learn to maintain the original layout distribution of $Q^{\mathrm{s}}$ and render sufficiently exaggerated caricatures following the shape conditions. Specifically, we use a layout energy function $\mathcal{E}_{\mathrm{layout}}$ to perform alignment between $Q=[q_{1}, q_{2},...,q_{n}]$ and $Q^{\mathrm{s}}=[q_{1}^{\mathrm{s}}, q_{2}^{\mathrm{s}},...,q_{n}^{\mathrm{s}}]$: 
\begin{small}
\begin{equation}
\mathcal{E}_{\mathrm{layout}} = \sqrt{\sum_{j=1}^{n} \| q_{j}-q_{j}^{\mathrm{s}}\|_{2}^{2} \cdot c^{\mathrm{s}}_{j}},
\end{equation}
\end{small}
where $c^{\mathrm{s}}_{j}\in[0,1]$ denotes a confidence weight indicating how strongly each token is associated with the sketch strokes. To compute weights, we first rasterize $S$ into a binary stroke map, where pixels on sketch strokes are assigned value $1$ and background pixels $0$. We then resize this map to the spatial resolution of the query features to obtain $C$. This selectively preserves the shape cues tied to informative sketch strokes while suppressing irrelevant regions.
 
If the face shapes of two generated caricatures are closely matched, they should localize textual condition signals similarly in cross-attention blocks. The attention map $\mathrm{Attn}(K_{\mathrm{txt}}, Q^{\mathrm{s}}, Q^{\mathrm{s}})$ of query $Q$ to the textual key $K_{txt}$ is hence used as additional signals to guide $\mathcal{P}^{\mathrm{i+s}}$ with semantic information:
\begin{equation}
\mathcal{E}_{\mathrm{sem}} = 
\|\mathrm{Attn}(K_{\mathrm{txt}}, Q, Q) - \mathrm{Attn}(K_{\mathrm{txt}}, Q^{\mathrm{s}}, Q^{\mathrm{s}})\|_{2}.
\end{equation}
The combined shape energy is
\begin{equation}
\mathcal{E}_{\mathrm{shape}} = \mathcal{E}_{\mathrm{layout}}+\mathcal{E}_{\mathrm{sem}}.
\end{equation}

\subsection{ID Alignment Guidance}
\label{sec:id}

Solely following the shape guidance from $\mathcal{P}^\mathrm{s}$ can lead to a significant degradation of ID fidelity. To improve ID consistency, previous works typically employ specialized loss functions \cite{Arcface, PhotoVerse, Portraitbooth}, which compare ID features between generated and reference images using feature extractors trained solely on real-domain photographs. However, the exaggerated shapes of caricatures introduce a significant domain gap, thereby reducing the effectiveness of these methods. Additionally, the ID alignment guidance should be performed across intermediate time-steps, where the predicted images are often noisy and flawed.

To overcome the issue, we design a novel ID energy function tailored for caricature synthesis. We guide $\mathcal{P}^{\mathrm{i+s}}$ using the ID-conditioned cross-attention outputs from $\mathcal{P}^{\mathrm{i}}$. Let
$O=[o_{1},...,o_{n}]$ and $O^{\mathrm{i}}=[o_{1}^{\mathrm{i}},...,o_{n}^{\mathrm{i}}]$ be the outputs of cross-attention layers, where $O^{\mathrm{i}}$ serves as the target.
However, direct alignment is inefficient, since $O$ may present a significantly different layout distribution compared to $O^{\mathrm{i}}$ when conditioned on the exaggerated shape $C_{\mathrm{s}}$.
Using $O^{\mathrm{i}}$ to guide $O$ naively will lead to a layout mismatch and confuse the model.
Therefore, we build associations based on the relevance between query tokens.
For each $o_{j}$, we find its corresponding $o_{k}^{\mathrm{i}}$ through
\begin{small}
\begin{equation}
k=\arg\max_{l}\Phi(q_{j}, q^{\mathrm{i}}_{l}),
\end{equation}
\end{small}
where $\Phi(\cdot,\cdot)$ is the cosine similarity  between two vectors.
The query-output pairs from $\mathcal{P}^{\mathrm{i}}$ behave like a ground-truth dictionary, where query tokens are ``keys'', and output tokens are ``values''. When a query token $q_{j}$ arrives, we refer to the dictionary, look up $q_{k}^{\mathrm{i}}$ that best matches $q_{j}$, and retrieve its corresponding output token $o_{k}^{\mathrm{i}}$ to provide guidance.

The ID energy is computed with a confidence weight $c^{\mathrm{i}}_{j}\in [0, 1]$ as
\begin{small}
\begin{equation}
\label{equation_l_id_refined}
\mathcal{E}_{\mathrm{id}}=\sqrt{\sum_{j=1}^{n}\|o_{j}-o_{k}^{\mathrm{i}}\|_{2}^{2}\cdot c^{\mathrm{i}}_{j}},
\end{equation}
\end{small}
The confidence weight $c^{\mathrm{i}}_{j}$ aims to adapt the guidance strength:
\begin{equation}
c^{\mathrm{i}}_{j}=\frac{\Phi(q_{j},q_{k}^{\mathrm{i}})-\phi}{1-\phi},\quad
\phi=\min_{l}\Phi(q_{j},q^{\mathrm{i}}_{l}).
\end{equation}
It helps weaken the guidance strength when the level of match between $q_{j}$ and $q_{k}^{\mathrm{i}}$ is not distinguishable from the pool of target query tokens.

\begin{figure*}[!htbp]
    \centering
    \includegraphics[width=\linewidth]{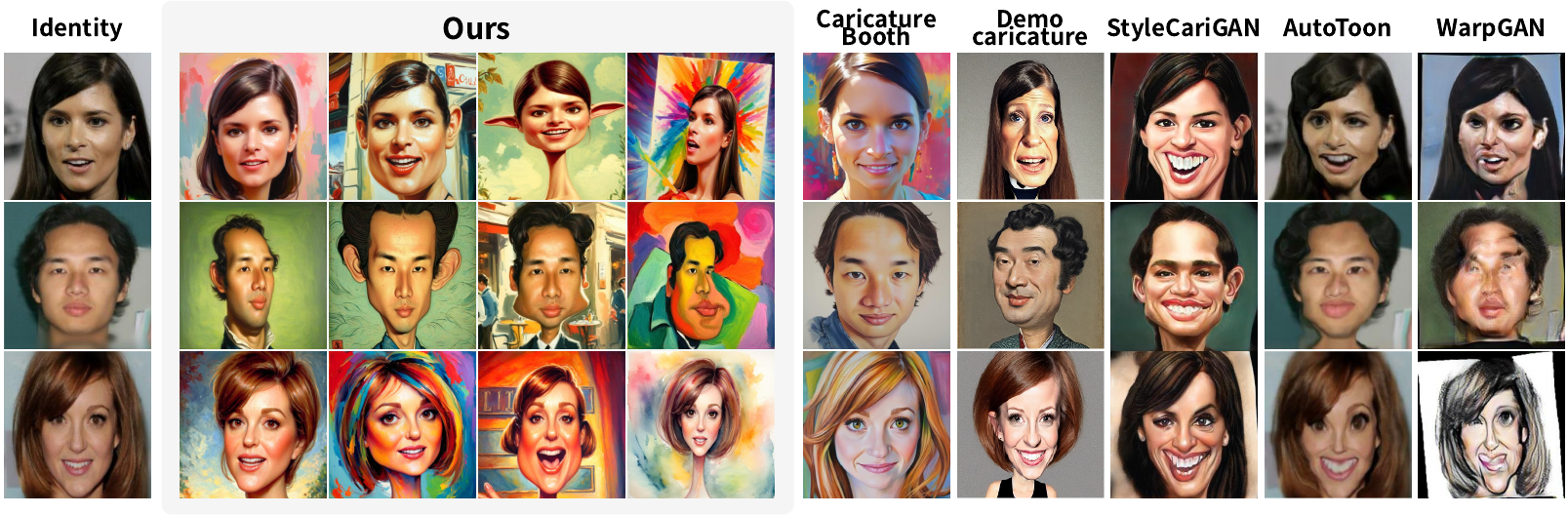}
    \vspace{-7mm}
    \caption{Qualitative performance comparison against caricature synthesis models.}
    \label{fig:comparison}
    \vspace{-3mm}
\end{figure*}

\subsection{Gradient Guidance with Balance Conditions}
$\mathcal{E}_\mathrm{shape}$ and $\mathcal{E}_\mathrm{id}$ are in opposition to each other. When the noisy latent is biased toward either side, the guidance strength from the other side increases, pulling the latent back to a balanced point. When combining both, the compositional energy function provides additional balance conditions $C_{\mathrm{b}}$ and helps the model find a sweet point on harmonizing ID and shape conditions with explicit gradient guidance. 
 
Let the existing conditions be $C_{\mathrm{e}}=[C_{\mathrm{txt}},C_{\mathrm{shape}},C_{\mathrm{id}}]$. The score function in this work can be expressed as: 
\begin{align}
&\nabla_{\hat{z}_{t}} \log q(\hat{z}_{t}|C_{\mathrm{b}},C_{\mathrm{e}})=
\nabla_{\hat{z}_{t}} \log \frac{q(\hat{z}_{t}|C_{\mathrm{e}})q(C_{\mathrm{b}}|\hat{z}_{t},C_{\mathrm{e}})}{q(C_{\mathrm{b}}|C_{\mathrm{e}})}\\
&= \nabla_{\hat{z}_{t}}\log q(\hat{z}_{t}|C_{\mathrm{e}}) + \nabla_{\hat{z}_{t}}\log q(C_{\mathrm{b}}|\hat{z}_{t},C_{\mathrm{e}}).
\end{align}

We apply classifier-free guidance \cite{CFG} to the first term and compute gradients of
$\mathcal{E}_{\mathrm{b}}=\mathcal{E}_{\mathrm{shape}}+\mathcal{E}_{\mathrm{id}}$ for the second. The final guided noise prediction $\tilde{\epsilon}_{t}$ is:
\begin{align}
\hat{\epsilon}_{t} &\gets (1+\gamma)\epsilon_{\theta}(\hat{z}_{t},t,C_{\mathrm{e}})-\gamma\epsilon_{\theta}(\hat{z}_{t},t,\emptyset), \\
\tilde{\epsilon}_{t} &\gets \hat{\epsilon}_{t} + \eta \nabla_{\hat{z}_{t}}\mathcal{E}_{\mathrm{b}}(\hat{z}_{t},t,C_{\mathrm{b}},C_{\mathrm{e}}),
\end{align}
where $\gamma$ controls classifier-free guidance and $\eta$ controls balance strength.

\subsection{Timestep-Constrained Guidance}
$\mathcal{E}_{\mathrm{shape}}$ guides layout distribution \cite{pulid} at a coarse scale, while $\mathcal{E}_{\mathrm{id}}$ guides detailed ID information at a relatively fine-grained scale. To fit the coarse-to-fine nature of the denoising process \cite{localize}, we only activate $\mathcal{E}_{\mathrm{shape}}$ and $\mathcal{E}_{\mathrm{id}}$ when $t$ falls in $[t^{\mathrm{s}}_{\mathrm{start}}, t^{\mathrm{s}}_{\mathrm{end}}]$ and $[t^{\mathrm{i}}_{\mathrm{start}}, t^{\mathrm{i}}_{\mathrm{end}}]$. We encourage the guidance of $\mathcal{E}_{\mathrm{shape}}$ at early time intervals, setting $t^{\mathrm{s}}_{\mathrm{start}}=T=1000$ and $t^{\mathrm{s}}_{\mathrm{end}}=700$. $\mathcal{E}_{\mathrm{id}}$ starts relatively late at $t^{\mathrm{i}}_{\mathrm{start}}=900$ and ends at $t^{\mathrm{i}}_{\mathrm{end}}=400$ to prevent harming the quality of results fine-grained details \cite{DragonDiffusion}.

\section{Experiments}
\label{sec:experiments}

\textbf{Implementation Details.}
Our model is built on SDXL \cite{SDXL} with the weights of Juggernaut-XL-v9 \cite{Juggernaut}. We use a pretrained ID Encoder from PuLID v1.1 \cite{pulid} and T2I-Sketch-Adapter-SDXL \cite{t2iadapter} to inject ID and shape conditions. During inference, the model is given textual prompts: ``A highly exaggerated and detailed caricature of a man/woman.''  The output image resolution is $768\times 768$. We use DPM++ 2M sampler \cite{DPM++2M} to accelerate inference speed, setting the number of inference steps $N$ to 40 and using a classifier-free guidance scale $\gamma$ of 7. The guidance rate $\eta$ is set to 0.4. The experiments are performed on a single RTX4090, and the code is implemented with PyTorch. All the other settings follow the original PuLID-ID-Encoder and T2I-Sketch-Adapter. 

\noindent\textbf{Dataset.}
We use the WebCaricature dataset \cite{WebCaricature} to evaluate the performance of different models. Specifically, we consistently select the photo with the least sequence number for each identity as the reference ID image $I$ and extract the edge map from each caricature associated with the identity as the sketch image $S$. The total number of samples is 1216. 

\noindent\textbf{Metrics.}
We use CLIP \cite{CLIP} scores to evaluate shape and ID consistency \cite{DemoCaricature}: S-CLIP measures shape consistency by comparing edge maps of generated and ground-truth caricatures, while I-CLIP measures identity consistency between generated caricatures and identity images. To assess overall quality, we further compare our model with the other caricature synthesis models using ImageReward \cite{ImageReward} and PickScore \cite{PickScore} as evaluation metrics.

\noindent\textbf{Baselines.}
The models for comparison include warping-based networks StyleCariGAN \cite{StyleCariGAN}, WarpGAN \cite{WarpGAN}, and AutoToon \cite{AutoToon}, also diffusion-based approaches DemoCaricature \cite{DemoCaricature}, and CaricatureBooth \cite{CaricatureBooth}.

\subsection{Qualitative Evaluation}
\Cref{fig:comparison} illustrates the performance of our model against several caricature synthesis models. StyleCariGAN \cite{StyleCariGAN}, WarpGAN \cite{WarpGAN}, and AutoToon \cite{AutoToon} cannot explicitly control the shape distortion. DemoCaricature \cite{DemoCaricature}, similar to this work, introduces explicit shape control by leveraging free-hand sketches. CaricatureBooth \cite{CaricatureBooth} guides the shape distortion with a fixed number of Bezier curves. Compared with the baselines, our model strikes a balance between preserving the facial features from identity images and ensuring a high level of creativity.

\Cref{fig:comparison_diffusion} evaluates our method against several state-of-the-art diffusion-based models. DemoCaricature \cite{DemoCaricature} is the first diffusion-based model tailored for caricature synthesis. PuLID \cite{pulid} is a tuning-free personalization model. We plug in a T2I-Sketch-Adapter \cite{t2iadapter} to provide additional shape control. Our model can generate high-quality caricatures in 16 seconds, which is approximately 4$\times$ faster than DemoCaricature without per-sample tuning. Compared to PuLID, our results align substantially better with the provided shape conditions after balancing the shape and ID information.

\subsection{Quantitative Evaluation}
It is worth mentioning that CaricatureBooth \cite{CaricatureBooth} only accepts a fixed number of Bezier curves as input and cannot be applied to free-hand sketches and edge maps, which makes it incompatible with the experiment. The results are shown in \Cref{tab:quantitative}.

\begin{figure}[!tbp]
    \centering
    \includegraphics[width=\linewidth]{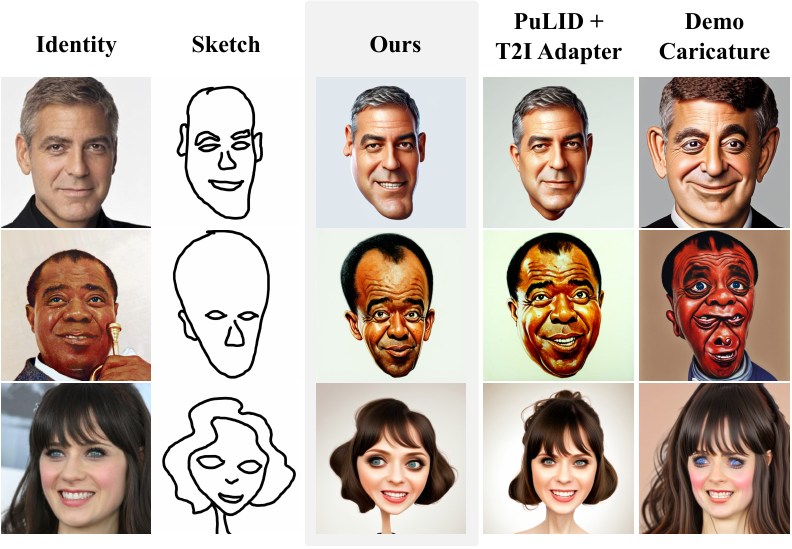}
     \vspace{-7mm}
    \caption{Qualitative performance comparison against diffusion-based models.}
    \label{fig:comparison_diffusion}
    \vspace{-3mm}
\end{figure}

When $\mathcal{E}_{\mathrm{shape}}$ and $\mathcal{E}_{\mathrm{id}}$ are disabled in turn, the model is biased toward ID preservation and shape exaggeration, and achieves the highest I-CLIP and S-CLIP score, respectively. By combining both, our final model finds a balanced point and generates results that respect both conditions (see \Cref{fig:losses}). As discussed before, custom evaluation models, including CLIP \cite{CLIP}, cannot extract accurate facial features of caricatures because of the domain gap, causing I-CLIP to be highly sensitive to the level of shape distortion. Although the final model gets a slightly lower I-CLIP score compared to DemoCaricature, the results still show that our model can sufficiently preserve ID information under highly creative shape conditions (see \Cref{fig:comparison_diffusion}). Additionally, all our variants consistently outperform counterparts in ImageReward and PickScore due to SDXL's strong generative capabilities, while their similar scores confirm that our gradient guidance preserves aesthetic quality.

\begin{table}[!h]
    \scriptsize
    \centering
    \setlength\tabcolsep{5.5pt}
    \begin{tabular}{ccccc}
    \toprule
       Methods          & I-CLIP $\uparrow$ & S-CLIP $\uparrow$ & ImageReward $\uparrow$ & PickScore $\uparrow$ \\
    \midrule
    StyleCariGAN& 0.5228    &   -  & 0.4340 & 0.0637 \\
    WarpGAN & 0.6634  & - &-0.2588& 0.1033\\
    AutoToon & \underline{0.7628} & -&-0.4978& 0.1094 \\
    DemoCaricature & 0.7591 & 0.8450 &0.2871&0.0949\\
    \arrayrulecolor{gray}\midrule\arrayrulecolor{black}
    Ours (full)    & 0.7512& \underline{0.8615} &\underline{0.8509}&\underline{0.2049}\\
    w/o $\mathcal{E}_{\mathrm{shape}}$ &  \textbf{0.7747}  &  0.8296 & \textbf{0.8967} & 0.2038 \\
    w/o $\mathcal{E}_{\mathrm{id}}$ &  0.7381  &  \textbf{0.8698} & 0.8443 & \textbf{0.2199} \\
    \bottomrule
    \end{tabular}
    \vspace{-2mm}
    \caption{Quantitative Comparison.}
    \label{tab:quantitative}
    \vspace{-3mm}
\end{table}

\subsubsection{User study}
We conduct a user study to evaluate the model's performance in terms of: (1) identity consistency with identity images, (2) shape consistency with sketches, and (3) overall quality. The study involved 20 identities (10 men and 10 women) and 20 exaggerated hand-drawn sketches (10 for men and 10 for women), resulting in a pool of 200 caricature samples. Each of the 16 volunteers received 10 samples, each asking them to rate the caricatures generated by different models on a discrete scale from 1 to 10 across the three aspects. The results are presented in \Cref{table:userstudy}. Our model achieves the highest ID consistency score, benefiting from humans' ability to capture facial features under shape deformations. By addressing the conflict with shape conditions, it also achieves the highest shape score, demonstrating overall superior performance.

\begin{table}[!t]
    \centering
    \setlength\tabcolsep{10pt}
    \begin{tabular}{cccc}
    \toprule
       Methods          & ID $\uparrow$& Shape $\uparrow$ & Overall $\uparrow$ \\
    \midrule
    StyleCariGAN& 4.59 &   -   &  4.91    \\
    WarpGAN & 4.81&- &4.62 \\
    AutoToon & 6.73 & - & 5.50\\
    DemoCaricature & 6.03& 5.51 & 6.06  \\
    \arrayrulecolor{gray}\midrule\arrayrulecolor{black}
    Ours     & \textbf{6.83}& \textbf{8.08}& \textbf{7.81} \\
    \bottomrule
    \end{tabular}
    \vspace{-1mm}
    \caption{Results of the user study. Each value represents the model's average score under a specified criterion.}
    \label{table:userstudy}
    \vspace{-3mm}
\end{table}

\begin{figure*}[!tbp]
    \centering
    \includegraphics[width=\linewidth]{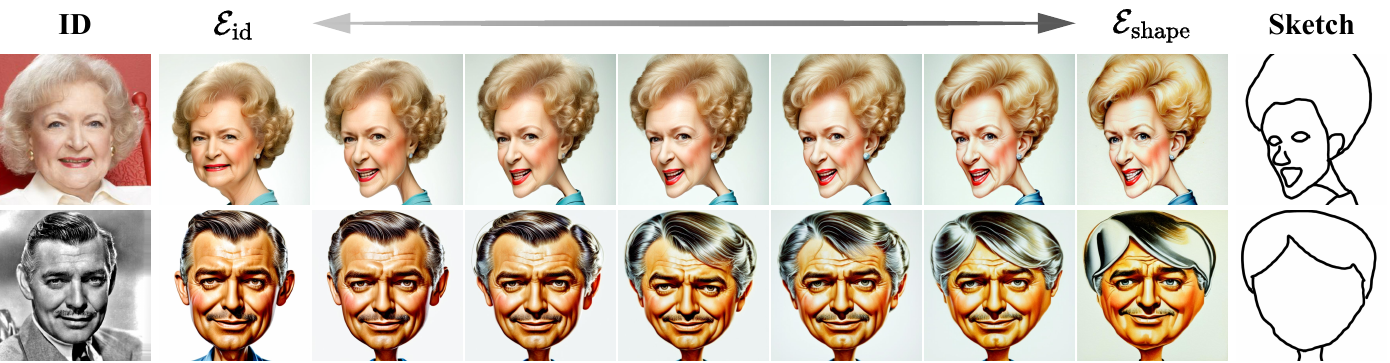}
     \vspace{-5mm}
    \caption{Interpolation of the scaling factors of $\mathcal{E}_{\mathrm{id}}$ and $\mathcal{E}_{\mathrm{shape}}$. From left to right, the scaling factor of $\mathcal{E}_{\mathrm{id}}$ linearly decreases from 2 to 0, while that of $\mathcal{E}_{\mathrm{shape}}$ linearly increases from 0 to 2. The results in the middle match our default setting.}
    \label{fig:transition}
    \vspace{-3mm}
\end{figure*}

\section{Ablation Study}

\begin{figure}[!htbp]
    \centering
    \includegraphics[width=\linewidth]{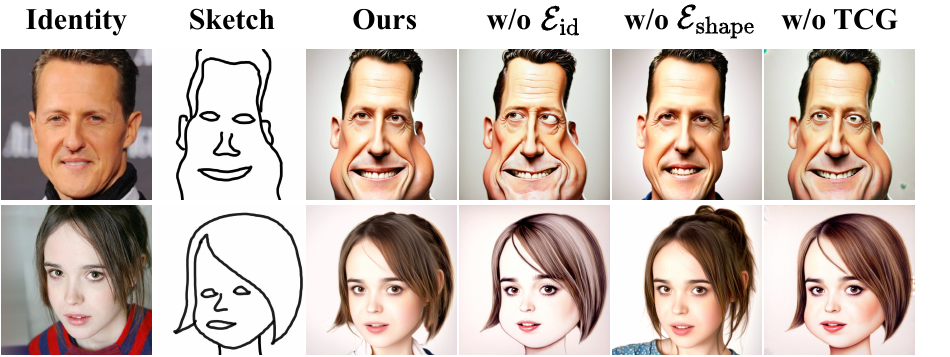}
    \vspace{-5mm}
    \caption{We disable different guidance terms respectively to demonstrate their functionalities. When timestep-constrained guidance (TCG) is disabled, $\mathcal{E}_{\mathrm{shape}}$ and $\mathcal{E}_{\mathrm{id}}$ are activated across all timesteps.}
    \label{fig:losses}
    \vspace{-3mm}
\end{figure}

\subsection{Impact of Different Guidance Terms}
\Cref{fig:losses} shows the results under different gradient guidance settings. Without $\mathcal{E}_{\mathrm{shape}}$, the model preserves facial structure well but fails to introduce sufficient exaggeration and lacks proper alignment with the target shape conditions, resulting in outputs that appear more like conventional portraits rather than expressive caricatures. Without $\mathcal{E}_{\mathrm{id}}$, the model is biased toward the shape conditions without respecting ID information, leading to outputs with diminished recognizability and less fine-grained textural details. In the absence of TCG, the model produces suboptimal results with less fine-grained details and generates unwanted artifacts that compromise the overall image quality. With the full version incorporating all guidance terms, the model achieves a balance where the generated caricatures faithfully adhere to the shape conditions while effectively preserving the identity-specific characteristics, demonstrating the competing yet complementary nature of each energy function in resolving the ID-shape conflict.

\subsection{Impact of Different Guidance Scales}
\Cref{fig:transition} presents the outputs produced when $\mathcal{E_{\mathrm{id}}}$ and $\mathcal{E_{\mathrm{shape}}}$ are weighted by different scaling factors. Increasing the scaling factor of $\mathcal{E}_{\mathrm{id}}$ while decreasing that of $\mathcal{E}_{\mathrm{shape}}$ yields results that preserve more identity information while following shape conditions less strictly, and vice versa. This smooth transition between identity-dominant and shape-dominant modes enables users to adjust the balance according to their specific preferences, demonstrating the flexibility of our model. 

\label{sec:extreme}
\begin{figure}[!htbp]
    \centering
    \includegraphics[width=\linewidth]{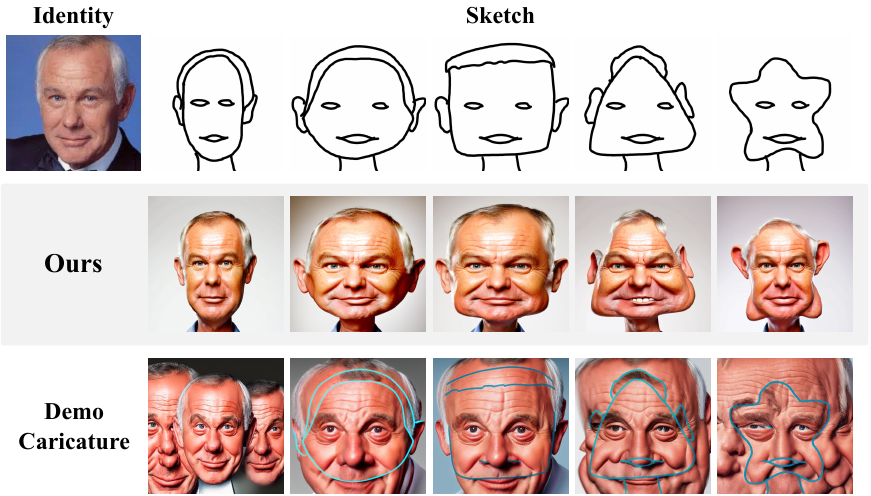}
    \vspace{-5mm}
    \caption{Performance comparison when sketches with different levels of exaggeration are used.}
    \label{fig:extreme}
    \vspace{-3mm}
\end{figure}

\subsection{Extreme Shape Exaggerations}
Our model shows robust compatibility with extreme shape conditions. As illustrated in \Cref{fig:extreme}, it can effectively resolve the conflict between ID preservation and shape exaggeration, producing caricatures with high ID fidelity even under highly exaggerated shape conditions. Notably, our approach maintains stable performance and generates high-quality outputs with recognizable identity characteristics even when the facial structures undergo drastic deformations. In contrast, the prior work \cite{DemoCaricature} struggles to interpret the spatial information from sketches when they are highly exaggerated because of the interruption of the conditioning signals. More examples are provided in Appendix B.

\section{Conclusion}
We present a caricature synthesis model that achieves a remarkable performance improvement by explicitly handling the ID-shape conflict. The core of our approach is a training-free framework that aligns hidden features in the cross-attention mechanism with those from two contrastive paths, enabling dynamic control over identity preservation and shape exaggeration. Tailored energy functions are designed to provide the model with additional information about the degree of deviation from either ID or shape conditions, pushing it to strike a balance on harmonizing both. By strategically harnessing the manifold of pretrained T2I models, our approach circumvents the data scarcity problem that hinders the development of caricature synthesis. The resulting framework possesses the capability to output high-quality caricatures that can not only preserve high ID fidelity but also retain the shape exaggeration defined by free-hand sketches. 

\clearpage
{
    \small
    \bibliographystyle{ieeenat_fullname}
    \bibliography{main}
}


  \input{supp}

\end{document}

%% file: supp.tex




%



\onecolumn
\setcounter{page}{1}
\section*{\centering
  \normalfont Supplementary Material for\\[0.3em]
  \bfseries CaricHarmony: Contrastive Diffusion Paths for Identity-Preserving Caricature Synthesis}
\section*{A. Algorithm}
The steps of our proposed method are summarized in Algorithm \ref{alg:balance} for clarity. We use DDIM sampler as an example for simplicity, but it is straightforward to replace it with the DPM++ 2M sampler \cite{DPM++2M}. $\bar\alpha_{t}=\prod_{i=1}^{t} \alpha_{i}$ is the cumulative product of the noise schedule \cite{DDPM}. It is worth mentioning that, at each time-step, the energy functions should be computed across all the cross-attention blocks and subsequently summed. 
\begin{algorithm}
  \caption{CaricHarmony}
  \label{alg:balance}
  \begin{algorithmic}[0]
    \STATE \textbf{Require:} UNET denoiser $\epsilon_{\theta}(\cdot)$; CLIP-T model $\mathcal{T}(\cdot)$; Word embedding layer $\mathcal{W}(\cdot)$; Text prompts $p$; Binary sketch image $S$; Identity image $I$; Classifier-free guidance scale $\gamma$; Guidance rate $\eta$; VAE decoder $D(\cdot)$.
    
    \STATE \textbf{Initialization:}
    \STATE Sample latent noise $\hat{z}_{T} = \hat{z}_{T}^{\mathrm{i}} = \hat{z}_{T}^{\mathrm{s}} \sim \mathcal{N}(0, \mathbf{I})$.
    \STATE Obtain text conditions $C_{\mathrm{txt}} \gets \mathcal{T}(\mathcal{W}(p))$.
    \STATE Obtain ID conditions $C_{\mathrm{id}} \gets PuLID(I)$.
    \STATE Obtain shape conditions $C_{\mathrm{s}} \gets Adapter(S)$.
    \STATE Set joint condition $C_{\mathrm{e}} \gets [C_{\mathrm{txt}}, C_{\mathrm{id}}, C_{\mathrm{s}}]$.
    
    \FOR{$t = T$ \TO $0$}
      \STATE $\mathcal{E}_{\mathrm{b}} \gets 0$.
      \STATE $\hat\epsilon_{t} \gets (1+\gamma)\,\epsilon_{\theta}(\hat{z}_{t}, t, C_{\mathrm{e}}) - \gamma\,\epsilon_{\theta}(\hat{z}_{t}, t, \emptyset)$
      \COMMENT {Classifier-Free Guidance} 
      \STATE Cache intermediate features $Q$, $K_{\mathrm{txt}}$, $V$, and $O$ derived from $\epsilon_{\theta}(\hat{z}_{t}, t, C_{\mathrm{e}})$.
      
      \IF{$t_{\mathrm{start}}^{\mathrm{s}} > t > t_{\mathrm{end}}^{\mathrm{s}}$}
        \STATE $\hat{z}_{t-1}^{\mathrm{s}}, Q^{\mathrm{s}} \gets DDIMSamplingStep(\hat{z}_{t}^{\mathrm{s}}, C_{\mathrm{s}}, C_{\mathrm{txt}}, \epsilon_{\theta}(\cdot), \gamma)$
        \COMMENT {A sampling step with Classifier-Free Guidance}
        \STATE $\mathcal{E}_{\mathrm{layout}}, \mathcal{E}_{\mathrm{sem}} \gets ComputeShapeEnergy(Q, Q^{\mathrm{s}}, K_{\mathrm{txt}})$
        \COMMENT {Equation (3) and Equation (4)} 
        \STATE $\mathcal{E}_{\mathrm{b}} \gets \mathcal{E}_{\mathrm{b}} + \mathcal{E}_{\mathrm{layout}} + \mathcal{E}_{\mathrm{sem}}$
      \ENDIF

      \IF {$t_{\mathrm{start}}^{\mathrm{i}} > t > t_{\mathrm{end}}^{\mathrm{i}}$}
        \STATE $\hat{z}_{t-1}^{\mathrm{i}}, [Q^{\mathrm{i}}, O^{\mathrm{i}}] \gets DDIMSamplingStep(\hat{z}_{t}^{\mathrm{i}}, C_{\mathrm{id}}, C_{\mathrm{txt}}, \epsilon_{\theta}(\cdot), \gamma)$
        \STATE $\mathcal{E}_{\mathrm{id}} \gets ComputeIDEnergy(Q, O, Q^{\mathrm{i}}, O^{\mathrm{i}})$
        \COMMENT {Equation (7)} 
        \STATE $\mathcal{E}_{\mathrm{b}} \gets \mathcal{E}_{\mathrm{b}} + \mathcal{E}_{\mathrm{id}}$
      \ENDIF

      \IF {$\mathcal{E}_{\mathrm{b}} \neq 0$}
        \STATE $\tilde\epsilon_{t} \gets \hat\epsilon_{t} + \eta \nabla_{\hat{z}_{t}}\mathcal{E}_{\mathrm{b}}$
      \ELSE
        \STATE $\tilde\epsilon_{t} \gets \hat\epsilon_{t}$
      \ENDIF
     
      \STATE $\hat{z}_{t-1} \gets \sqrt{\bar\alpha_{t-1}} \frac{\hat{z}_{t} - \sqrt{1-\bar\alpha_{t}}\,\tilde\epsilon_{t}}{\sqrt{\bar\alpha_{t}}} + \sqrt{1-\bar\alpha_{t-1}}\,\tilde{\epsilon}_{t}$
      \COMMENT {DDIM Sampling}
    \ENDFOR
    \STATE $\hat{x}_{0} \gets D(\hat{z}_{0})$
    \RETURN $\hat{x}_{0}$
  \end{algorithmic}
\end{algorithm}

\section*{B. Results in Various Levels of Exaggerations}
\begin{figure*}[!htbp]
    \centering
    \includegraphics[width=\linewidth]{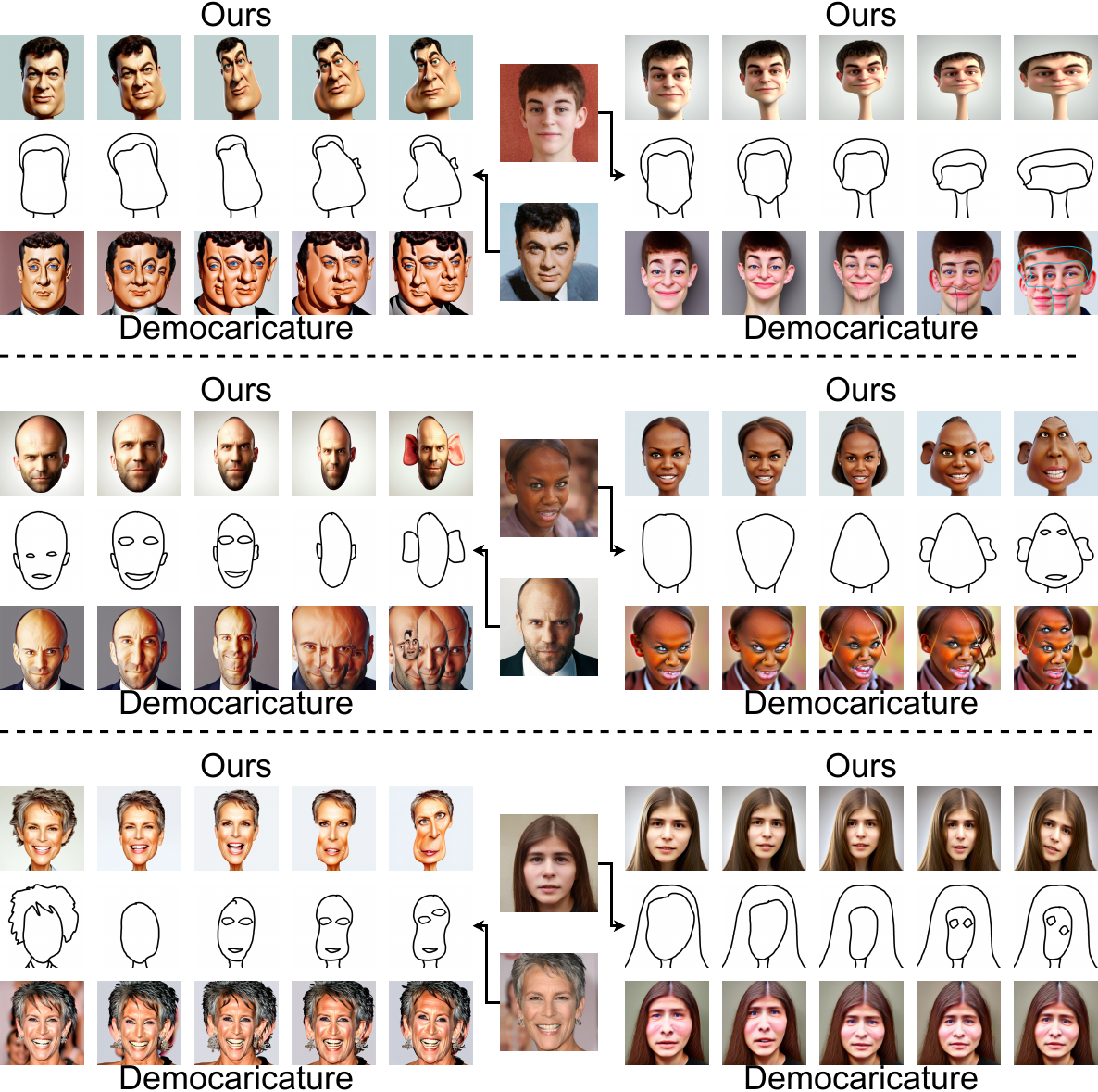}
    \caption{Additional qualitative performance comparison on sketches with varying levels of exaggeration.}
    \label{fig:extreme_extra}
\end{figure*}

Figure \ref{fig:extreme_extra} presents additional results to demonstrate the robustness and compatibility of our model when handling a diverse range of shape conditions. The results are organized into three groups from top to bottom. In each group, the left side takes a celebrity photo as the reference identity image, while the right side takes a synthetically generated face. Even when the sketches are highly exaggerated, the model successfully preserves the intended creative semantics while maintaining strong ID fidelity, highlighting its effectiveness in balancing shape and ID conditions.

\section*{C. Additional Results}
\begin{figure*}[!htbp]
    \centering
    \includegraphics[width=\linewidth]{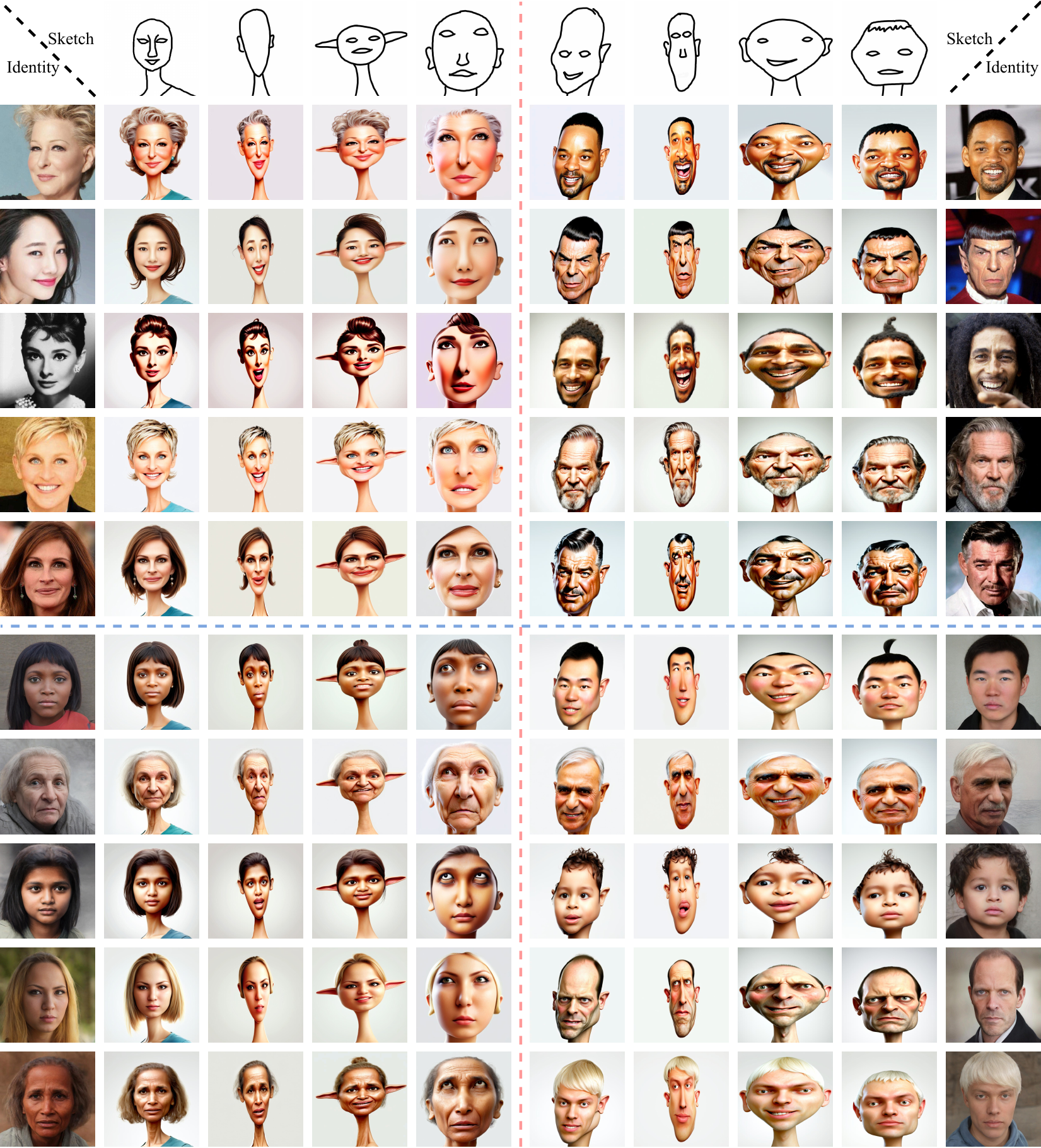}
    \caption{Additional results of our model with various identity images and shape conditions. Left of the pink line: female faces. Right of the pink line: male faces. Above the blue line: celebrity faces. Below the blue line: unknown faces.}
    \label{fig:matrix}
\end{figure*}

 We present additional results with different combinations of ID and shape conditions and build a comprehensive result matrix as shown in Figure \ref{fig:matrix}. Each cell in the matrix represents a unique pairing of a specific identity and distinct shape conditions. By observing rows and columns where either ID or shape conditions remain fixed while the other varies, it can be seen that the model can preserve ID consistency across shape changes and maintain shape characteristics across different identities. The results highlight the strong robustness of our model in disentangling and recombining ID and shape conditions. Thanks to the robust generalization capability of the PuLID-ID-encoder \cite{pulid}, the model still retains such ability for unknown faces.
 
\section*{D. Failure Cases}
\begin{figure*}[!htbp]
    \centering
    \includegraphics[width=\linewidth]{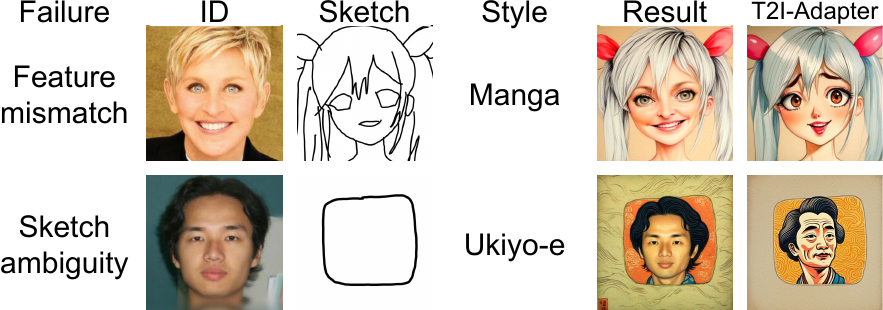}
    \caption{Failure cases of our model. Results under the column T2I-Adapter are those generated from $\mathcal{P}^{\mathrm{s}}$.}
    \label{fig:failure}
\end{figure*}

We discuss the failure cases of our model as shown in Figure \ref{fig:failure}. One issue arises when the features presented in the sketch are significantly different from those in the reference identity: the model will not ignore but signify them in the generated results, causing degradation of ID fidelity. For example, in the first row of Figure \ref{fig:failure}, the sketch depicts a character with pigtails, whereas the identity has short golden hair. To handle the contradiction, the model tends to overly follow the shape conditions (i.e., the pigtails), while also incorporating irrelevant features (such as hair color) from $\mathcal{P}^{\mathrm{s}}$ into the output. This leads to an attribute shift (a change of hair color from golden to white) that harms ID fidelity. Another issue stems from our model's dependency on the T2I-Sketch-Adapter \cite{t2iadapter}. Suppose the sketch is highly ambiguous and cannot be interpreted by the original T2I-Sketch-Adapter correctly. In that case, our model will follow inappropriate guidance from $\mathcal{P}^{\mathrm{s}}$ and produce undesired results. In the second row of Figure \ref{fig:failure}, we expect the model to generate a caricature with a square-shaped face. However, the T2I-Sketch-Adapter misinterprets the shape conditions as a backdrop, leading to failure to produce the desired result.

\section*{E. Additional Comparison against Modern Models}
\begin{figure}[!tbp]
    \centering
    \includegraphics[width=\linewidth]{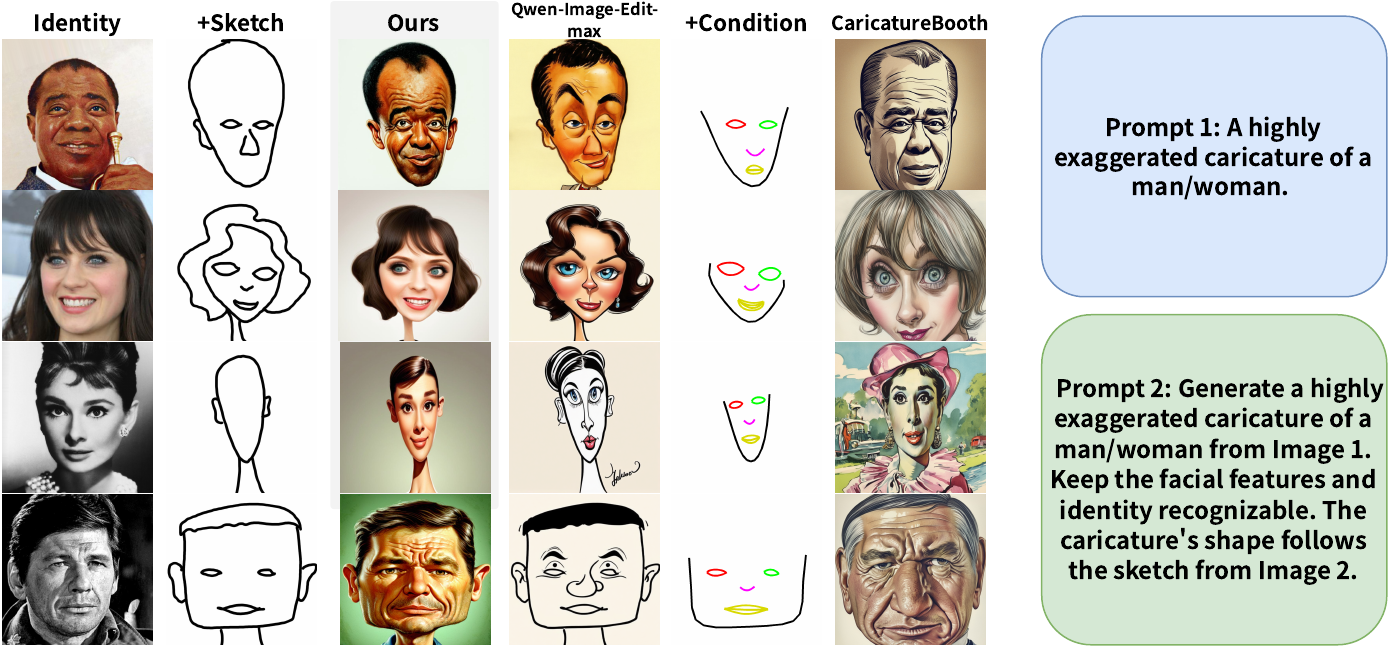}
    \vspace{-5mm}
    \caption{Performance comparison against modern generative models. We use Prompt 1 for our model and CaricatureBooth, and Prompt 2 for Qwen-Image-Edit-Max to fit the image editing task.}
    \vspace{-3mm}
    \label{fig:comparison_latest}
\end{figure}

We further compare our model with CaricatureBooth \cite{CaricatureBooth} and QWEN-Image-Edit-Max\footnote{
\url{https://www.alibabacloud.com/help/en/model-studio/qwen-image-edit-api}} \cite{Qwen} in Figure \ref{fig:comparison_latest}. CaricatureBooth is the latest caricature synthesis model based on SDXL \cite{SDXL}. QWEN-Image-Edit-Max is a 20B multi-modal model for high-fidelity image editing that supports multi-image input, allowing us to provide both the identity image and sketch condition simultaneously. We annotate hand-drawn sketches as Bezier curves representing the chin, lips, eyes, and nose to make them compatible with CaricatureBooth. However, CaricatureBooth cannot sufficiently capture sketch shapes due to its restrictive input format with a fixed number of control points, resulting in poor representation of additional facial features and highly curved regions. QWEN-Image-Edit-Max, on the other hand, tends to overly follow the sketch shape while neglecting the subject's identity. Our model can effectively harmonize both conditions via the introduced balancing mechanism while supporting flexible sketch formats. We use the same data used in the user study for quantitative evaluation. As shown in Table \ref{tab:quantitative_latest}, our model achieves the best ID and shape consistency score, as well as the best aesthetic score according to ImageReward. 

\begin{table}[!htbp]
    \centering
    \begin{tabular}{ccccc}
    \toprule
       Methods          & I-CLIP $\uparrow$ & S-CLIP $\uparrow$ & PickScore $\uparrow$ & ImageReward $\uparrow$ \\
    \midrule
    Qwen-Image-Edit-Max & 0.5487    &   0.8374  & 0.1274 & -0.3983 \\
    CaricatureBooth & 0.5615  & 0.7637 & \textbf{0.5141}& 0.4827\\
    \arrayrulecolor{gray}\midrule\arrayrulecolor{black}
    Ours   & \textbf{0.6472} & \textbf{0.8500} &0.3584&\textbf{0.8493}\\
    \bottomrule
    \end{tabular}
    \vspace{-1mm}
    \caption{Quantitative comparison with concurrent models.}
    \label{tab:quantitative_latest}
    \vspace{-3mm}
\end{table}

\section*{F. Face Reconstruction Capability}
To evaluate the identity preserving capability of our model, we perform face reconstruction for all identities in WebCaricature \cite{WebCaricature}, using the lowest-indexed photo as the ID image and extracting its edge map or landmarks as shape conditions. We use the cosine similarity of ArcFace \cite{Arcface} features as an additional metric. As shown in Table \ref{tab:face_reconstruction}, our method achieves a significantly higher ArcFace similarity and a competitive I-CLIP score compared to modern caricature synthesis models, demonstrating that our method can effectively preserve facial features of identities.

\begin{table}
    \centering
    \begin{tabular}{cccc}
    \toprule
       Methods         & I-CLIP $\uparrow$& ArcFace-sim $\uparrow$ \\
    \midrule
    DemoCaricature & \textbf{0.8182}& 0.5405\\
    CaricatureBooth  & 0.6450  & 0.4989  \\
    \arrayrulecolor{gray}\midrule\arrayrulecolor{black}
    Ours & 0.8121  &   \textbf{0.6929}\\
    \bottomrule
    \end{tabular}
    \vspace{-1mm}
    \caption{Results of face reconstruction.}
    \label{tab:face_reconstruction}
    \vspace{-3mm}
\end{table}


%% file: main.bib
@String(IJCV = {Int. J. Comput. Vis.})

@String(CVPR= {IEEE Conf. Comput. Vis. Pattern Recog.})

@String(ICCV= {Int. Conf. Comput. Vis.})

@String(BMVC= {Brit. Mach. Vis. Conf.})

@String(TOG= {ACM Trans. Graph.})

@String(TMM  = {IEEE Trans. Multimedia})

@String(ICLR = {Int. Conf. Learn. Represent.})

@String(AAAI = {AAAI})

@String(IJCV  = {IJCV})

@String(CVPR  = {CVPR})

@String(ICCV  = {ICCV})

@String(BMVC  =	{BMVC})

@String(TOG   = {ACM TOG})

@String(TMM   =	{IEEE TMM})

@String(ICLR  = {ICLR})

@inproceedings{pulid,
  title={Pulid: Pure and lightning id customization via contrastive alignment},
  author={Guo, Zinan and Wu, Yanze and Zhuowei, Chen and Zhang, Peng and He, Qian and others},
  booktitle={NeurIPS},
  year={2024}
}

@inproceedings{t2iadapter,
  title={T2i-adapter: Learning adapters to dig out more controllable ability for text-to-image diffusion models},
  author={Mou, Chong and Wang, Xintao and Xie, Liangbin and Wu, Yanze and Zhang, Jian and Qi, Zhongang and Shan, Ying},
  booktitle={AAAI},
  year={2024}
}

@article{CariGAN,
  title={Carigans: Unpaired photo-to-caricature translation},
  author={Cao, Kaidi and Liao, Jing and Yuan, Lu},
  journal={arXiv preprint arXiv:1811.00222},
  year={2018}
}

@inproceedings{WarpGAN,
  title={Warpgan: Automatic caricature generation},
  author={Shi, Yichun and Deb, Debayan and Jain, Anil K},
  booktitle={CVPR},
  year={2019}
}

@article{CariMe,
  title={Carime: Unpaired caricature generation with multiple exaggerations},
  author={Gu, Zheng and Dong, Chuanqi and Huo, Jing and Li, Wenbin and Gao, Yang},
  journal={IEEE TMM},
  year={2021}
}

@article{StyleCariGAN,
  title={Stylecarigan: caricature generation via stylegan feature map modulation},
  author={Jang, Wonjong and Ju, Gwangjin and Jung, Yucheol and Yang, Jiaolong and Tong, Xin and Lee, Seungyong},
  journal={ACM TOG},
  year={2021},
}

@inproceedings{SENet,
  title={Squeeze-and-excitation networks},
  author={Hu, Jie and Shen, Li and Sun, Gang},
  booktitle={CVPR},
  year={2018}
}

@inproceedings{AutoToon,
  title={Autotoon: Automatic geometric warping for face cartoon generation},
  author={Gong, Julia and Hold-Geoffroy, Yannick and Lu, Jingwan},
  booktitle={WACV},
  year={2020}
}

@inproceedings{StyleGAN,
  title={A style-based generator architecture for generative adversarial networks},
  author={Karras, Tero and Laine, Samuli and Aila, Timo},
  booktitle={CVPR},
  year={2019}
}

@article{Semantic-CariGAN,
  title={Learning to caricature via semantic shape transform},
  author={Chu, Wenqing and Hung, Wei-Chih and Tsai, Yi-Hsuan and Chang, Yu-Ting and Li, Yijun and Cai, Deng and Yang, Ming-Hsuan},
  journal={IJCV},
  year={2021}
}

@inproceedings{DemoCaricature,
  title={Democaricature: Democratising caricature generation with a rough sketch},
  author={Chen, Dar-Yen and Bhunia, Ayan Kumar and Koley, Subhadeep and Sain, Aneeshan and Chowdhury, Pinaki Nath and Song, Yi-Zhe},
  booktitle={CVPR},
  year={2024}
}

@article{DDPM,
  title={Denoising Diffusion Probabilistic Models},
  author={Ho, Jonathan and Jain, Ajay and Abbeel, Pieter},
  journal={NeurIPS},
  year={2020}
}

@inproceedings{SD,
  title={High-resolution image synthesis with latent diffusion models},
  author={Rombach, Robin and Blattmann, Andreas and Lorenz, Dominik and Esser, Patrick and Ommer, Bj{\"o}rn},
  booktitle={CVPR},
  year={2022}
}

@inproceedings{Imagen,
  title={Photorealistic text-to-image diffusion models with deep language understanding},
  author={Saharia, Chitwan and Chan, William and Saxena, Saurabh and Li, Lala and Whang, Jay and Denton, Emily L and Ghasemipour, Kamyar and Gontijo Lopes, Raphael and Karagol Ayan, Burcu and Salimans, Tim and others},
  booktitle={NeurIPS},
  year={2022}
}

@article{IP-Adapter,
  title={IP-Adapter: Text Compatible Image Prompt Adapter for Text-to-Image Diffusion Models},
  author={Ye, Hu and Zhang, Jun and Liu, Sibo and Han, Xiao and Yang, Wei},
  journal={arXiv preprint arxiv:2308.06721},
  year={2023}
}

@inproceedings{SDXL,
  title={Sdxl: Improving latent diffusion models for high-resolution image synthesis},
  author={Podell, Dustin and English, Zion and Lacey, Kyle and Blattmann, Andreas and Dockhorn, Tim and M{\"u}ller, Jonas and Penna, Joe and Rombach, Robin},
  booktitle={ICLR},
  year={2024}
}

@inproceedings{Textual-Inversion,
  title={An image is worth one word: Personalizing text-to-image generation using textual inversion},
  author={Gal, Rinon and Alaluf, Yuval and Atzmon, Yuval and Patashnik, Or and Bermano, Amit H and Chechik, Gal and Cohen-Or, Daniel},
  booktitle={ICLR},
  year={2023}
}

@inproceedings{DreamBooth,
  title={Dreambooth: Fine tuning text-to-image diffusion models for subject-driven generation},
  author={Ruiz, Nataniel and Li, Yuanzhen and Jampani, Varun and Pritch, Yael and Rubinstein, Michael and Aberman, Kfir},
  booktitle={CVPR},
  year={2023}
}

@inproceedings{Custom-Diffusion,
  title={Multi-concept customization of text-to-image diffusion},
  author={Kumari, Nupur and Zhang, Bingliang and Zhang, Richard and Shechtman, Eli and Zhu, Jun-Yan},
  booktitle={CVPR},
  year={2023}
}

@inproceedings{Perfusion,
  title={Key-locked rank one editing for text-to-image personalization},
  author={Tewel, Yoad and Gal, Rinon and Chechik, Gal and Atzmon, Yuval},
  booktitle={ACM SIGGRAPH},
  year={2023}
}

@inproceedings{InstantBooth,
  title={Instantbooth: Personalized text-to-image generation without test-time finetuning},
  author={Shi, Jing and Xiong, Wei and Lin, Zhe and Jung, Hyun Joon},
  booktitle={CVPR},
  year={2024}
}

@inproceedings{Arcface,
  title={Arcface: Additive angular margin loss for deep face recognition},
  author={Deng, Jiankang and Guo, Jia and Xue, Niannan and Zafeiriou, Stefanos},
  booktitle={CVPR},
  year={2019}
}

@article{PhotoVerse,
  title={Photoverse: Tuning-free image customization with text-to-image diffusion models},
  author={Chen, Li and Zhao, Mengyi and Liu, Yiheng and Ding, Mingxu and Song, Yangyang and Wang, Shizun and Wang, Xu and Yang, Hao and Liu, Jing and Du, Kang and others},
  journal={arXiv preprint arXiv:2309.05793},
  year={2023}
}

@inproceedings{Portraitbooth,
  title={Portraitbooth: A versatile portrait model for fast identity-preserved personalization},
  author={Peng, Xu and Zhu, Junwei and Jiang, Boyuan and Tai, Ying and Luo, Donghao and Zhang, Jiangning and Lin, Wei and Jin, Taisong and Wang, Chengjie and Ji, Rongrong},
  booktitle={CVPR},
  year={2024}
}

@inproceedings{localize,
  title={Localizing object-level shape variations with text-to-image diffusion models},
  author={Patashnik, Or and Garibi, Daniel and Azuri, Idan and Averbuch-Elor, Hadar and Cohen-Or, Daniel},
  booktitle={ICCV},
  year={2023}
}

@inproceedings{WebCaricature,
  author    = {Jing Huo and
               Wenbin Li and
               Yinghuan Shi and
               Yang Gao and
               Hujun Yin},
  title     = {WebCaricature: a benchmark for caricature recognition},
  booktitle = {BMVC},
  year      = {2018}
}

@inproceedings{Transformer,
  title={Attention is all you need},
  author={Vaswani, Ashish and Shazeer, Noam and Parmar, Niki and Uszkoreit, Jakob and Jones, Llion and Gomez, Aidan N and Kaiser, {\L}ukasz and Polosukhin, Illia},
  booktitle={NeurIPS},
  year={2017}
}

@inproceedings{residual,
  title={Deep residual learning for image recognition},
  author={He, Kaiming and Zhang, Xiangyu and Ren, Shaoqing and Sun, Jian},
  booktitle={CVPR},
  year={2016}
}

@inproceedings{CLIP,
  title={Learning transferable visual models from natural language supervision},
  author={Radford, Alec and Kim, Jong Wook and Hallacy, Chris and Ramesh, Aditya and Goh, Gabriel and Agarwal, Sandhini and Sastry, Girish and Askell, Amanda and Mishkin, Pamela and Clark, Jack and others},
  booktitle={ICML},
  year={2021},
}

@inproceedings{UNET,
  title={U-net: Convolutional networks for biomedical image segmentation},
  author={Ronneberger, Olaf and Fischer, Philipp and Brox, Thomas},
  booktitle={MICCAI},
  year={2015}
}

@inproceedings{ImageReward,
  title={Imagereward: Learning and evaluating human preferences for text-to-image generation},
  author={Xu, Jiazheng and Liu, Xiao and Wu, Yuchen and Tong, Yuxuan and Li, Qinkai and Ding, Ming and Tang, Jie and Dong, Yuxiao},
  booktitle={NeurIPS},
  year={2023}
}

@inproceedings{PickScore,
  title={Pick-a-pic: An open dataset of user preferences for text-to-image generation},
  author={Kirstain, Yuval and Polyak, Adam and Singer, Uriel and Matiana, Shahbuland and Penna, Joe and Levy, Omer},
  booktitle={NeurIPS},
  year={2023}
}

@article{DPM++2M,
  title={Dpm-solver++: Fast solver for guided sampling of diffusion probabilistic models},
  author={Lu, Cheng and Zhou, Yuhao and Bao, Fan and Chen, Jianfei and Li, Chongxuan and Zhu, Jun},
  journal={Machine Intelligence Research},
  year={2025}
}

@inproceedings{CFG,
  title={Classifier-free diffusion guidance},
  author={Ho, Jonathan and Salimans, Tim},
  booktitle={NeurIPS Workshop},
  year={2021}
}

@inproceedings{score-based-model,
  title={Improved techniques for training score-based generative models},
  author={Song, Yang and Ermon, Stefano},
  booktitle={NeurIPS},
  year={2020}
}

@inproceedings{CG,
  title={Diffusion models beat gans on image synthesis},
  author={Dhariwal, Prafulla and Nichol, Alexander},
  booktitle={NeurIPS},
  year={2021}
}

@inproceedings{sketch-guided-generation,
  title={Sketch-guided text-to-image diffusion models},
  author={Voynov, Andrey and Aberman, Kfir and Cohen-Or, Daniel},
  booktitle={ACM SIGGRAPH},
  year={2023}
}

@inproceedings{mask-guided-generation,
  title={High-fidelity guided image synthesis with latent diffusion models},
  author={Singh, Jaskirat and Gould, Stephen and Zheng, Liang},
  booktitle={CVPR},
  year={2023},
}

@inproceedings{score-based-image-editing,
  title={Diffusion self-guidance for controllable image generation},
  author={Epstein, Dave and Jabri, Allan and Poole, Ben and Efros, Alexei and Holynski, Aleksander},
  booktitle={NeurIPS},
  year={2023}
}

@inproceedings{DragonDiffusion,
  title={Dragondiffusion: Enabling drag-style manipulation on diffusion models},
  author={Mou, Chong and Wang, Xintao and Song, Jiechong and Shan, Ying and Zhang, Jian},
  booktitle={ICLR},
  year={2024}
}

@inproceedings{CaricatureBooth,
  title={CaricatureBooth: Data-Free Interactive Caricature Generation in a Photo Booth},
  author={Qu, Zhiyu and Miao, Yunqi and Zhang, Zhensong and Song, Jifei and Deng, Jiankang and Song, Yi-Zhe},
  booktitle={CVPR},
  year={2025}
}

@inproceedings{TPS,
  title={Splines minimizing rotation-invariant semi-norms in Sobolev spaces},
  author={Duchon, Jean},
  booktitle={Constructive theory of functions of several variables: Proceedings of a conference held at Oberwolfach April 25--May 1, 1976},
  year={2006},
}

@article{Bezier,
  title={Curve fitting with Bezier cubics},
  author={Shao, Lejun and Zhou, Hao},
  journal={Graphical models and image processing},
  year={1996},
}

@inproceedings{
  Diffusion-SDE,
  title={Score-Based Generative Modeling through Stochastic Differential Equations},
  author={Yang Song and Jascha Sohl-Dickstein and Diederik P Kingma and Abhishek Kumar and Stefano Ermon and Ben Poole},
  booktitle={ICLR},
  year={2021},
}

@inproceedings{Langevin,
  title={Deep unsupervised learning using nonequilibrium thermodynamics},
  author={Sohl-Dickstein, Jascha and Weiss, Eric and Maheswaranathan, Niru and Ganguli, Surya},
  booktitle={ICML},
  year={2015},
}

@inproceedings{Song-Langevin,
  title={Generative modeling by estimating gradients of the data distribution},
  author={Song, Yang and Ermon, Stefano},
  booktitle={NeurIPS},
  year={2019}
}

@misc{Juggernaut,
  title        = "{Juggernaut XL v9 + RunDiffusion Photo v2}",
  author       = "{RunDiffusion}",
  year         = {2024}, 
  howpublished={\url{https://huggingface.co/RunDiffusion/Juggernaut-XL-v9}}
}

@inproceedings{LatexBlend,
  title={LatexBlend: Scaling multi-concept customized generation with latent textual blending},
  author={Jin, Jian and Yu, Zhenbo and Shen, Yang and Fu, Zhenyong and Yang, Jian},
  booktitle={CVPR},
  year={2025}
}

@article{Imagine-yourself,
  title={Imagine yourself: Tuning-free personalized image generation},
  author={He, Zecheng and Sun, Bo and Juefei-Xu, Felix and Ma, Haoyu and Ramchandani, Ankit and Cheung, Vincent and Shah, Siddharth and Kalia, Anmol and Subramanyam, Harihar and Zareian, Alireza and others},
  journal={arXiv preprint arXiv:2409.13346},
  year={2024}
}

@article{flux-already-knows,
  title={Flux Already Knows--Activating Subject-Driven Image Generation without Training},
  author={Kang, Hao and Fotiadis, Stathi and Jiang, Liming and Yan, Qing and Jia, Yumin and Liu, Zichuan and Chong, Min Jin and Lu, Xin},
  journal={arXiv preprint arXiv:2504.11478},
  year={2025}
}

@misc{flux,
  title        = "{Flux.1-dev}",
  author       = "{Black Forest Labs}",
  year         = {2024}, 
  howpublished={\url{https://github.com/black-forest-labs/flux}},
}

@inproceedings{LoRACLR,
  title={LoRACLR: Contrastive Adaptation for Customization of Diffusion Models},
  author={Simsar, Enis and Hofmann, Thomas and Tombari, Federico and Yanardag, Pinar},
  booktitle={CVPR},
  year={2025}
}

@article{Qwen,
  title={Qwen-image technical report},
  author={Wu, Chenfei and Li, Jiahao and Zhou, Jingren and Lin, Junyang and Gao, Kaiyuan and Yan, Kun and Yin, Sheng-ming and Bai, Shuai and Xu, Xiao and Chen, Yilei and others},
  journal={arXiv preprint arXiv:2508.02324},
  year={2025}
}
